\documentclass[letterpaper, 10 pt, conference]{ieeeconf}  

\IEEEoverridecommandlockouts                              

\usepackage{siunitx}
\usepackage[dvipsnames]{xcolor}
\usepackage{colortbl}
\definecolor{green}{rgb}{0.5, 1.0, 0.83}

\usepackage{graphicx}
\usepackage{booktabs}
\usepackage{multirow}
\usepackage{subcaption}
\usepackage{bm}
\usepackage{amsmath, amssymb, latexsym}
\usepackage[acronym]{glossaries}
\usepackage{pbox}
\newacronym{rl}{RL}{reinforcement learning}
\newacronym{marl}{MARL}{multi-agent reinforcement learning}
\newacronym{decpomdp}{Dec-POMDP}{decentralized partially observable Markov decision process}
\newacronym{posg}{POSG}{partially observable stochastic game}
\newacronym{ppo}{PPO}{proximal policy optimization}
\newacronym{gae}{GAE}{generalized advantage estimation}
\newacronym{dof}{DoF}{degrees of freedom}
\newacronym{lstm}{LSTM}{long short-term memory}

\usepackage{acronym}
\acrodef{lstm}[LSTM]{Long Short-Term Memory}
\acrodefindefinite{lstm}{an}{a}

\DeclareMathOperator{\ReLU}{ReLU}

\usepackage{tikz}
\usetikzlibrary{backgrounds}
\usetikzlibrary{calc}
\usepackage{pgfplots}
\pgfplotsset{compat=1.16}
\usepgfplotslibrary{fillbetween}
\usepackage{xcolor}
\definecolor{conv}{HTML}{EAFBAD}
\definecolor{fc}{HTML}{00FF7F}
\definecolor{lstm}{HTML}{007FFF}
\tikzset{fc/.style={black,draw=black,fill=fc,rectangle,minimum height=0.6cm}}
\tikzset{conv/.style={black,draw=black,fill=conv,rectangle,minimum height=0.6cm}}
\tikzset{lstm/.style={black,draw=black,fill=lstm,rectangle,minimum height=0.6cm}}
 
\usepackage{xspace}

\title{\LARGE \bf
Cooperative Assistance in Robotic Surgery through\\ Multi-Agent Reinforcement Learning
}

\author{Paul Maria Scheikl$^{1\dag}$, Balázs Gyenes$^{1\dag}$, Tornike Davitashvili$^{2}$, Rayan Younis$^{2}$, André Schulze$^{2}$,\\ Beat P. Müller-Stich$^{2}$, Gerhard Neumann$^{1}$, Martin Wagner$^{2\ddag}$, and Franziska Mathis-Ullrich$^{1\ddag}$
\thanks{*The present contribution is supported by the Helmholtz Association under the joint research school "HIDSS4Health – Helmholtz Information and Data Science School for Health".}
\thanks{$^{\dag, \ddag}$ These authors contributed equally.}
\thanks{$^{1}$ P. M. Scheikl, B. Gyenes, G. Neumann, and F. Mathis-Ullrich are with the Institute for Anthropomatics and Robotics, Karlsruhe Institute of Technology, 76131 Karlsruhe, Germany. \newline {\small corresponding author: \tt franziska.ullrich@kit.edu}}%
\thanks{$^{2}$ T. Davitashvili, R. Younis, A. Schulze, B. P. Müller-Stich, and M. Wagner are with the Department for General, Visceral and Transplantation Surgery, Heidelberg University Hospital, 69120 Heidelberg, Germany.}%
}

\begin{document}

\maketitle
\thispagestyle{empty}
\pagestyle{empty}

\begin{abstract}
Cognitive cooperative assistance in robot-assisted surgery holds the potential to increase quality of care in minimally invasive interventions.
Automation of surgical tasks promises to reduce the mental exertion and fatigue of surgeons.
In this work, multi-agent reinforcement learning is demonstrated to be robust to the distribution shift introduced by pairing a learned policy with a human team member.
Multi-agent policies are trained directly from images in simulation to control multiple instruments in a sub task of the minimally invasive removal of the gallbladder.
These agents are evaluated individually and in cooperation with humans to demonstrate their suitability as autonomous assistants.
Compared to human teams, the hybrid teams with artificial agents perform better considering completion time (44.4\% to 71.2\% shorter) as well as number of collisions (44.7\% to 98.0\% fewer).
Path lengths, however, increase under control of an artificial agent (11.4\% to 33.5\% longer).
A multi-agent formulation of the learning problem was favored over a single-agent formulation on this surgical sub task, due to the sequential learning of the two instruments.
This approach may be extended to other tasks that are difficult to formulate within the standard reinforcement learning framework.
Multi-agent reinforcement learning may shift the paradigm of cognitive robotic surgery towards seamless cooperation between surgeons and assistive technologies.
\end{abstract}

  \tikz[remember picture, overlay]
    \node [rotate=0, text width=20cm] at ($(current page.north)+(0,-1)$)
      {© 2021 IEEE.  Personal use of this material is permitted. Permission from IEEE must be obtained for all other uses, in any current or future media, including reprinting/republishing this material for advertising or promotional purposes, creating new collective works, for resale or redistribution to servers or lists, or reuse of any copyrighted component of this work in other works.};

\section{INTRODUCTION}
\label{sec:introduction}
    Minimally invasive surgery of the abdomen, i.e. laparoscopy, usually cannot be performed by an individual surgeon alone.
    A team of at least two trained surgeons is required to operate three to five rod-shaped surgical instruments, including a laparoscopic camera.
    In robot-assisted laparoscopy, however, a multi-armed robot is usually teleoperated by a single surgeon~\cite{maeso2010efficacy}.
    The surgeon operates two instruments at a time via mechanical interfaces, while the remaining assistive instruments remain immobile.
    Thus, to adapt the camera view or reapply tension to grasped tissue, the surgeon must change instrument control and pause the actual operation, which takes time and may be mentally exhausting.
    As a result, one of the key criticisms of current robot-assisted laparoscopy is long operation times~\cite{maeso2010efficacy}.
    This challenge may be overcome by equipping the robot with autonomous behavior to cooperate with the surgeon by automatically operating surgical instruments.
    
    In contrast to rule-based methods and approaches from programming by demonstration, \gls{rl} learns autonomous behaviors from interactions with an environment.
    In the context of robot-assisted surgery, any \gls{rl} system must be robust to humans who may retake manual control of a variable number of instruments, thus changing the overall action space.
    For instance, masking out a portion of a single \gls{rl} agent's action space may lead to unexpected behavior, since it effectively changes the environment dynamics the agent has experienced during training.
    In contrast, each agent of a multi-agent system treats the other agents as part of its stochastic environment and trains against a diverse range of behaviors from them.
    This fluid team composition of humans and agents, which may be deployed at will to take control of individual robotic arms, motivates the use of decentralized policies.

    \begin{figure}[t]
         \centering
         \begin{subfigure}[b]{0.49\linewidth}
             \centering
             \includegraphics[trim={0cm 7cm 0cm 5cm},clip,width=\linewidth]{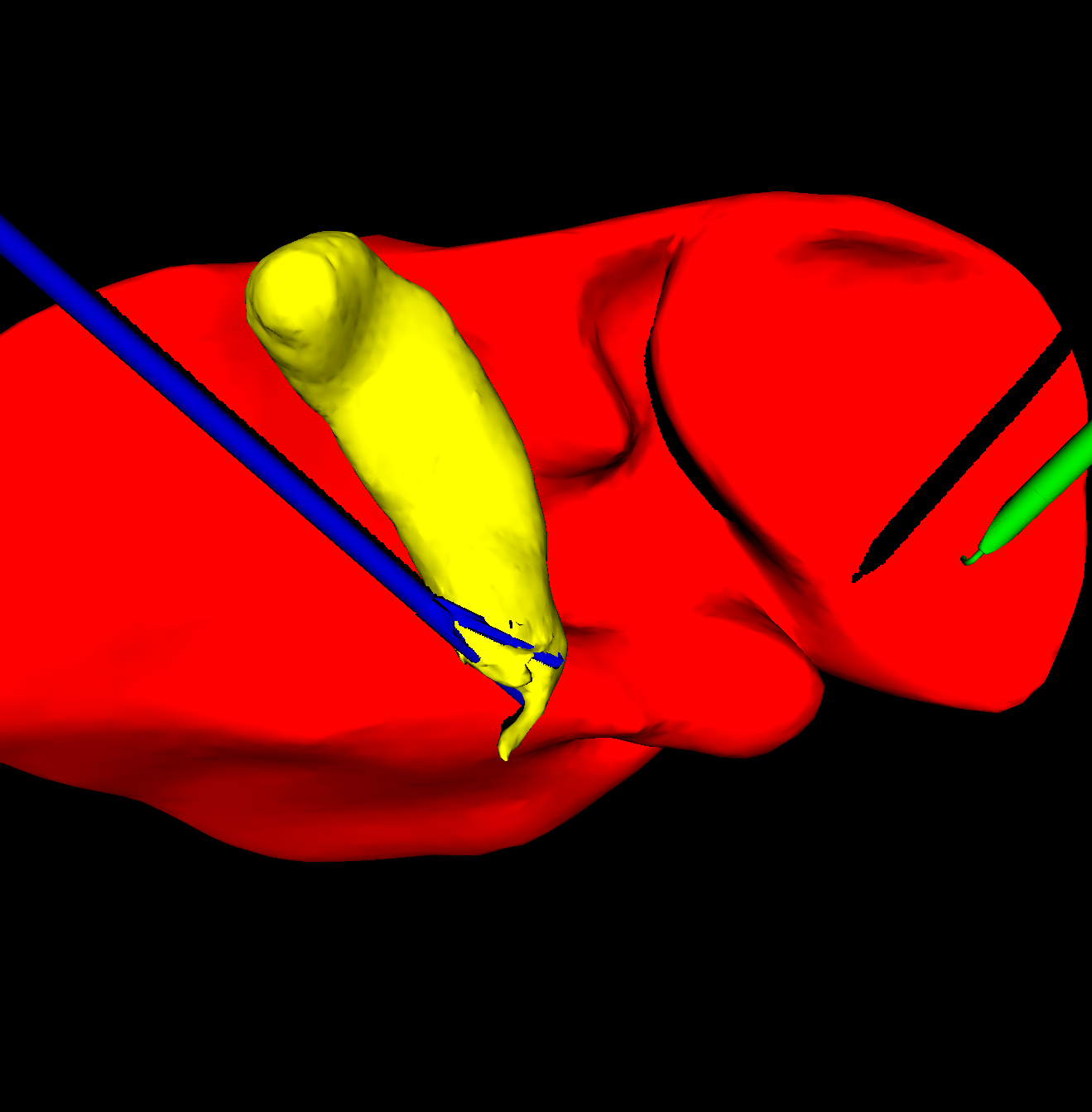}
             \label{fig:task_a}
         \end{subfigure}
         \hfill
         \begin{subfigure}[b]{0.49\linewidth}
             \centering
             \includegraphics[trim={0cm 7cm 0cm 5cm},clip,width=\linewidth]{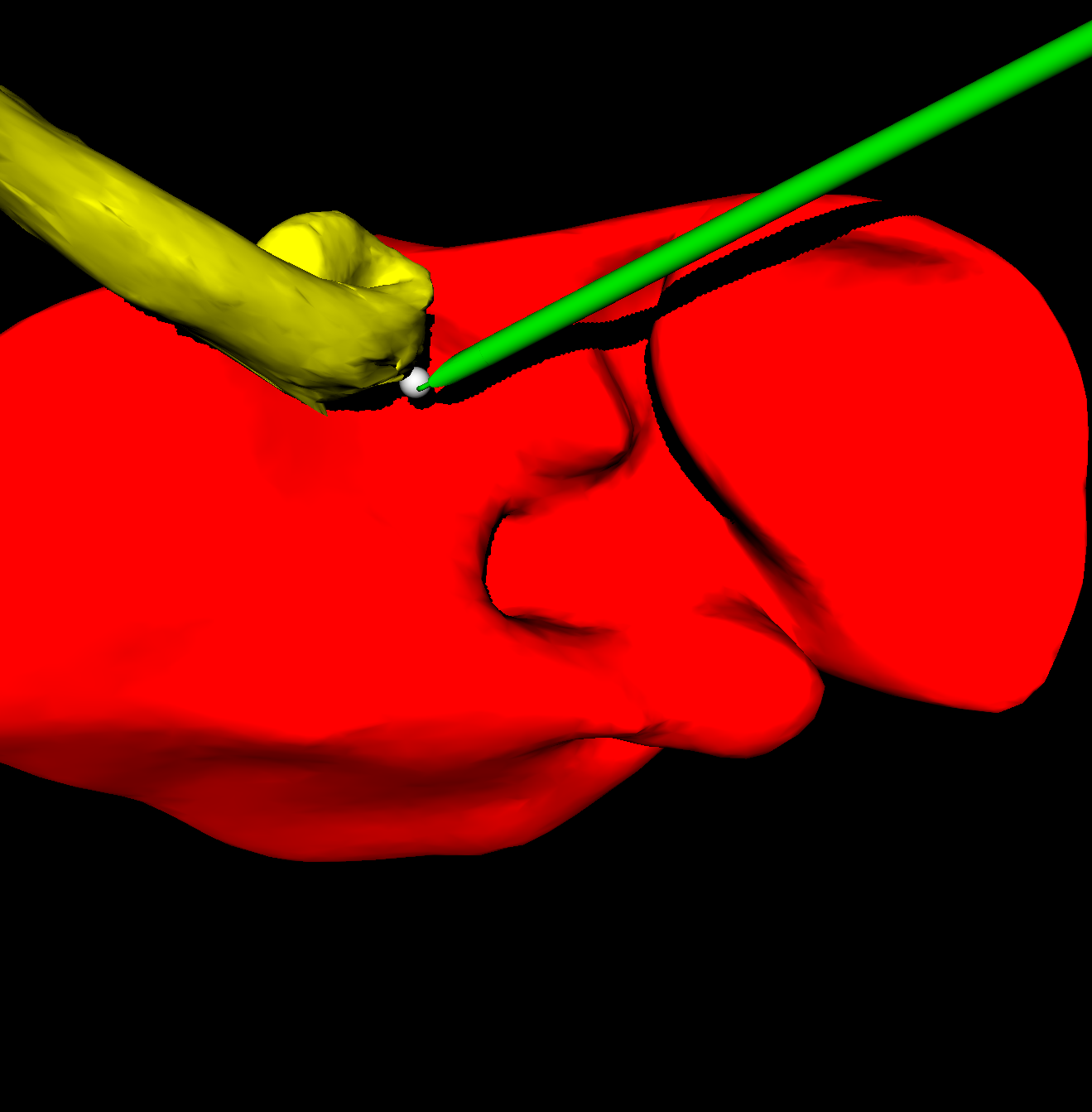}
             \label{fig:task_b}
         \end{subfigure}
         \vspace{-0.8cm}
         \caption{Sample task from laparoscopic cholecystectomy. To solve the task, the gripper (blue) grasps and lifts the gallbladder (yellow) to expose a visual target (white) so that the cauterization hook (green) may touch the visual target without colliding with liver (red) or gallbladder tissue.}
        \label{fig:task}
    \vspace{-0.5cm}
    \end{figure}

    In robot-assisted surgery, \gls{rl} has been applied to learn kinematics and dynamics of complex systems, such as concentric tubes~\cite{iyengarInvestigatingExplorationDeep2020b} and soft tissue~\cite{shinAutonomousTissueManipulation2019b} from interactions with an environment, as well as to directly learn policies for robotic control of surgical instruments.
    However, these applications are restricted to small state and action spaces that require precise registration of the scene~\cite{tagliabue2020soft, richter2019OpenSourcedRL}, visual markers~\cite{thananjeyanMultilateralSurgicalPattern2017b},~\cite{nguyen2019manipulating}, or intensive preprocessing of the sensory data~\cite{kellerOpticalCoherenceTomographyGuided2020a}.
    
    Extending the state of the art, this work aims to learn decentralized robotic policies for laparoscopic surgery directly from raw image data to control all degrees of freedom of several laparoscopic instruments in an environment with deformable objects.
    In contrast to prominent works in the field of \gls{marl}, such as learning cooperation in hide and seek environments~\cite{baker2019emergent} or computer games such as DOTA~2~\cite{OpenAI_dota} and StarCraft~2~\cite{rashid2018qmix}, the agents in laparoscopic surgery are very heterogeneous, as one instrument's task cannot be performed by another, e.g. the camera cannot cut tissue.
    However, all agents share a common observation of the scene, as the endoscopic camera is the sole source of information.
    This contribution presents the first work on training decentralized policies on cooperative tasks in robot-assisted surgery with reinforcement learning methods.

\section{METHODS}
\label{sec:methods}
  \subsection{Approach}
    The policies are trained in simulation without human interaction directly on image data from a simulated endoscopic camera.
    Policy performance is evaluated in (a) a fully autonomous setting to demonstrate that the agents are able to learn the cooperative task and (b) in hybrid team compositions with a trained agent and a human to demonstrate that the learned policies may be used as cooperative assistants.
    A single experienced human surgeon and a team of two human surgeons-in-training serve as the baseline for performance evaluation.

    A sub task from the minimally invasive removal of the gallbladder, i.e. laparoscopic cholecystectomy, serves as the scenario to evaluate our approach (see Fig.~\ref{fig:task}).
    However, the presented method is not limited to the chosen sub task and may be transferred to other cooperative tasks in robot-assisted surgery.

  \subsection{Multi-Agent Reinforcement Learning}
    \subsubsection{Partially Observable Stochastic Games}
        Laparoscopic surgery is a fully cooperative multi-agent task that can be described as a \gls{posg}~\cite{hansen2004dynamic}.
        Each agent is unaware of the other agents' actions and is able to choose from its own set of actions.
        Here, observations are shared among all agents, as images from a single laparoscopic camera are the sole source of information about the environment.
        A \gls{posg} consists of a tuple $G = \langle S, \boldsymbol{A}, P, R, Z, O, n, \gamma \rangle$.
        At each time step, $s \in S$ represents the ground truth state of the environment.
        Each agent $i \in \{ 1, \ldots, n \}$ selects an individual action $a_i \in A_i$ to form a joint action $\boldsymbol{a} \in \boldsymbol{A} \equiv \{ A_1 \times \ldots \times A_n \}$ that is applied to the environment.
        This results in a change of state according to the environment's state transition function $P(s'| s, \boldsymbol{a}): S \times \boldsymbol{A} \times S \rightarrow [0, 1]$.
        All agents share a common observation $z \in Z$ generated by an observation function $O(s): S \rightarrow Z$, but receive individual rewards drawn from reward function $R_i(s): S \rightarrow \mathbb{R}$.
        
        This work trains distributed recurrent policies, conditioned on each agent's individual action-observation history $\tau_i \in T_i \equiv (Z \times A_i)^{*}$.
        Actions are sampled from each agent's stochastic policy $\pi_i(a_i|\tau_i): T_i \times A_i \rightarrow [0, 1]$.
        Each agent is optimized to maximize its own discounted return $R_{i,t} = \sum_{j=0}^{\infty} \gamma^j r_{i,t+j}$ with discount factor $\gamma$.

    \subsubsection{Learning Environment}
        The learning environment represents a sub task from laparoscopic cholecystectomy, i.e. the removal of the gallbladder.
        During dissection of the gallbladder from the liver, a gripper lifts the gallbladder and maintains tension on it to expose the remaining connective tissue between gallbladder and liver.
        Subsequently, an electric cauterization hook (cauter) moves towards the next cutting point on the connective tissue.
        The environment is initiated with slightly randomized instrument positions and one of three different target positions on the border between gallbladder and liver.
        The target is visualized as a white sphere that is initially covered by the gallbladder, as illustrated in Fig.~\ref{fig:task}.

        \begin{figure}[tb]
             \centering
             \vspace{1.5mm}
             \includegraphics[trim={0cm 2cm 0cm 0cm},clip,width=\linewidth]{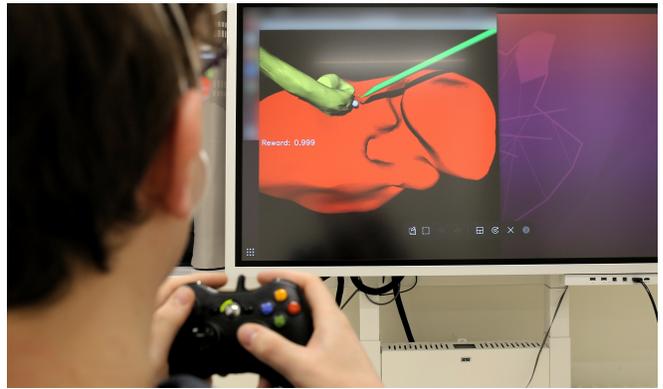}
             \caption{A human agent receives a $1024\times1024$ RGB image of the scene and controls the instruments in four degrees of freedom with an Xbox controller.}
             \label{fig:human_obs}
        \vspace{-0.5cm}
        \end{figure}

        The deformable physics simulation of the learning environment was implemented in the Simulation Open Framework Architecture (SOFA)~\cite{faure2012sofa}.
        The agents were trained using simulation time steps of $\frac{1}{30}$ \si{\second}, so they are well-matched with the humans interacting with the simulation that runs at \SI{30}{\hertz}, as shown in Fig.~\ref{fig:human_obs}.
        The deformable parts of the surgical scene are organ models of the liver and gallbladder extracted from anonymized human computed tomography scans and consist of finite element meshes of tetrahedral elements.
        The rigid laparoscopic instruments of the task are (1) a gripper that begins the episode having already grasped the gallbladder by its neck and (2) a cauter at a distance to the gallbladder.
        An episode is considered successful if the cauter tip reaches the target within a tolerance of \SI{5}{\mm}.
        Exceeding a time limit of 1000 steps (i.e., equivalent to approximately \SI{33}{\second}) is considered a loss, as is losing all grasping contacts between gripper and gallbladder due to excessive pulling.
        Regrasping is not considered in the scope of this work.
        The outcome of an episode is thus characterized by one of three events: 1) target successfully reached, 2) time run-out, and 3) lost grasp on the gallbladder.

    \subsubsection{Reward, State and Action Spaces}
    \label{sec:methods:spaces}
        Instrument movements in laparoscopic surgery are restricted by a trocar placed into the incision in the abdominal wall that serves as a pivot point.
        The complete six dimensional pose of a laparoscopic instrument can thus be described by four independent \gls{dof}, namely pan (pivoted left/right motion), tilt (pivoted up/down motion), spin (rotation around the instrument axis), and insertion depth (along the instrument axis) relative to the pivot point.

        The simulation environment conforms to the OpenAI Gym API~\cite{brockman2016gym}, applying an action at each time step to the environment and receiving an observation and reward value for the next time step, as well as a done signal.
        The action space of artificial agents is represented as a discrete choice over eight possible movement actions, two for each of the four independent \gls{dof} with a step size of $1^\circ$ of rotation or \SI{1}{\mm} insertion, and a no-op action for no movement.
        The shared observation for the artificial agents are RGB images with a resolution of $128\times128$ pixels.
        Human agents, however, receive a higher resolution image with $1024\times1024$ pixels, also showing the current reward on the left side of the screen, as illustrated in Fig.~\ref{fig:human_obs}.
        The human agents output actions in a continuous space, characterized by real displacement in each \gls{dof} ($a^i~\in~[-1, 1]^4$).
        The reward for the cauter at each time step consists of a weighted sum of (1) the Euclidean distance from the cauter to the target and (2) penalties for collisions between the cauter and other objects in the scene.
        The reward for the gripper consists of a weighted sum of (1) the number of visible pixels of the target, (2) the number of mesh elements of the gallbladder that obstruct the view of the target, (3) the number of lost grasp contacts, (4) the insertion depth of the gripper, and (5) penalties for collisions between the gripper and objects other than the gallbladder.
        Reaching the target with the cauter when the target is visible ends the episode and assigns a positive reward for completion to both agents.
        Losing the grasp also ends the episode, but assigns a negative reward to the gripper.

    \subsubsection{Independent PPO}
    Here, the reinforcement learning framework rlpyt~\cite{stooke2019rlpyt} and its implementation of \gls{ppo}~\cite{schulman2017proximal} were utilized to train the artificial agents.
    \gls{ppo} is a popular policy-gradient algorithm, which has been successfully applied to single-agent~\cite{tagliabue2020soft} and multi-agent~\cite{baker2019emergent} problems.
    \gls{ppo} is naturally extended to the multi-agent case by summing up policy loss due to each agent, using each agent's respective advantage estimation conditioned on its observation.
    Advantage was calculated using \gls{gae}~\cite{schulman2015high} based on the agent's state-value estimates.
    Value estimation for both agents was trained using mean squared error (MSE) loss against targets calculated by TD($\lambda$).
    Agents were trained with a discount of $\gamma =$ \num{0.99} and \gls{gae} $\lambda =$ \num{0.8}.
    Batch size was set to \num{2560} steps unrolled through time with learning rate $\alpha =$ \num{3e-4}.
    The \gls{ppo} clip ratio was \num{0.1}, with \num{4} minibatches per epoch and \num{4} epochs per iteration.
    Advantages were normalized over each iteration.
    A loss term proportional to the negative policy entropy was included, with coefficient \num{0.01}.
    Gradients were clipped to a maximum overall norm of \num{1.0}.
    
    \subsubsection{Agent Architectures}
      
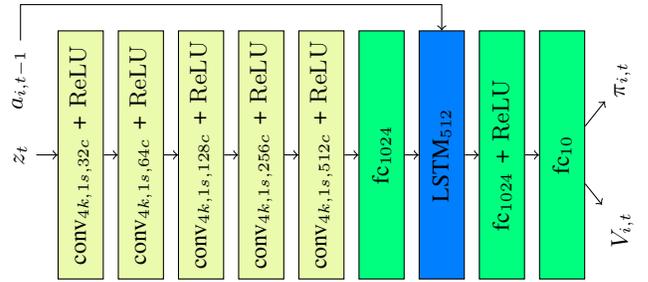
\begin{figure}[tb]
  \centering
  \vspace{1.5mm}
  \begin{tikzpicture}
    \node[rotate=90] (img) at (0,0) {\small$z_t$};

    \node[conv,rotate=90,minimum width=3.3cm] (conv1) at (0.8,0) {\small$\text{conv}_{4k, 1s, 32c}$ + $\ReLU$};
    \node[conv,rotate=90,minimum width=3.3cm] (conv2) at (1.6,0) {\small$\text{conv}_{4k, 1s, 64c}$ + $\ReLU$};
    \node[conv,rotate=90,minimum width=3.3cm] (conv3) at (2.4,0) {\small$\text{conv}_{4k, 1s, 128c}$ + $\ReLU$};
    \node[conv,rotate=90,minimum width=3.3cm] (conv4) at (3.2,0) {\small$\text{conv}_{4k, 1s, 256c}$ + $\ReLU$};
    \node[conv,rotate=90,minimum width=3.3cm] (conv5) at (4,0) {\small$\text{conv}_{4k, 1s, 512c}$ + $\ReLU$};
    \node[fc  ,rotate=90,minimum width=3.3cm] (fc1) at (4.8,0) {\small$\text{fc}_{1024}$};
    
    \node[rotate=90] (action) at (0,1) {\small $a_{i, t-1}$};
    
    \node[lstm,rotate=90,minimum width=3.3cm] (lstm) at (5.6,0) {\small$\text{LSTM}_{512}$};
    \node[fc  ,rotate=90,minimum width=3.3cm] (fc2) at (6.4,0) {\small$\text{fc}_{1024}$ + $\ReLU$};
    \node[fc  ,rotate=90,minimum width=3.3cm] (fc3) at (7.2,0) {\small$\text{fc}_{10}$};

    \node[rotate=90] (act) at (8,1) {\small$\pi_{i, t}$};
    \node[rotate=90] (val) at (8,-1) {\small$V_{i, t}$};

    \draw[->] (img) -- (conv1);
    \draw[->] (conv1) -- (conv2);
    \draw[->] (conv2) -- (conv3);

    \draw[->] (conv3) -- (conv4);
    \draw[->] (conv4) -- (conv5);

    \draw[->] (conv5) -- (fc1);
    \draw[->] (fc1) -- (lstm);

    \draw[->] (lstm) -- (fc2);
    \draw[->] (fc2) -- (fc3);

    \draw[->] (fc3) -- (act);
    \draw[->] (fc3) -- (val);
    
    \draw[->] (action.east) -- node {} (0, 2) -| (lstm.east);

  \end{tikzpicture}
  \caption{Illustration of the artificial agents' neural networks. Convolutional layers (kernel size $k$, stride $s$, channel $c$), followed by a combination of fully connected and LSTM layers. Outputs are logits ($\pi_{i, t}$) and value ($V_{i, t}$).}
  \label{fig:network}
  \vspace{-0.6cm}
\end{figure}

      The architecture illustrated in Fig.~\ref{fig:network} was used for both artificial agents, with no weight sharing between them.
      The neural network consists of five convolutional layers for encoding the image observations and a combination of fully-connected and \gls{lstm} layers to represent non-linear, time-dependent behaviors.
      The neural networks receive one RGB frame and the agent's previous action $a_{i,t-1}$ at a time and output logits $\pi_{i,t}$ for each action $a_{i,t}$ and a value estimate $V_{i,t}$ corresponding to agent $i$.
      A softmax function converts the logits to action probabilities.
      \vspace{-0.5mm}

  \subsection{Team Compositions}
  \label{sec:methods:teams}
    \begin{figure}[b]
        \centering
        \vspace{-2.5mm}
        \includegraphics[width=\linewidth]{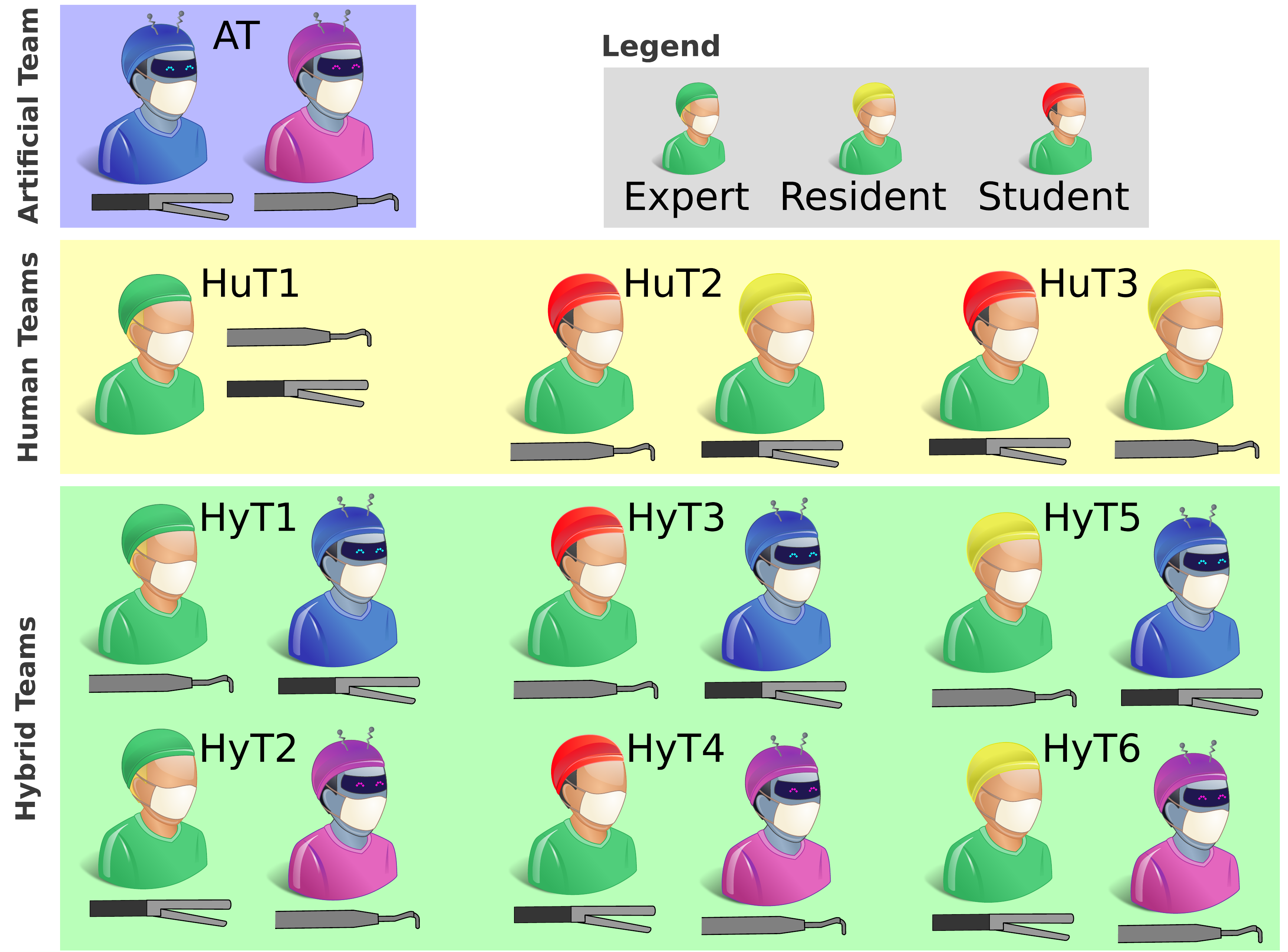}
        \caption{Composition of artificial (blue, AT), human (yellow, HuT1-3), and hybrid (green, HyT1-6) teams.}
        \label{fig:teams}
          \vspace{-1mm}
    \end{figure}

  The laparoscopic task was evaluated by one artificial team, three different human teams, and a total of six different hybrid teams, as illustrated in Fig.~\ref{fig:teams}.
  The artificial team (AT) consists of two decentralized policies that are trained to control one specific instrument each.
  The first human team (HuT1) consists of a single expert surgeon with six years of surgical experience who controls one instrument at a time. In contrast to all other teams in this experiment the expert may change the active instrument at will.
  The second human team (HuT2) consists of a surgical resident controlling the gripper, while a medical student controls the cauter.
  The third human team (HuT3) consists of the same surgical resident and medical student but with reversed instrument roles.
  The six hybrid teams (HyT1-6) each consist of one human agent (\textit{expert}, \textit{resident}, or \textit{student}) controlling either gripper or cauter and the respective artificial agent controlling the other instrument.

  \subsection{Human Interaction with the Environment}
    Humans interact with the simulation through Xbox controllers (Microsoft Corp., USA) polled with a frequency of \SI{30}{\hertz}.
    The \gls{dof} \textit{insert}, \textit{pan}, and \textit{tilt} are mapped to the analog sticks.
    The \textit{spin} is controlled through the shoulder buttons, but is not needed to complete the task.
    The single expert surgeon controls one instrument at a time and is able to change between instruments by pressing a dedicated button on the controller.
    Each member of the second and third human teams has dedicated control over one instrument.

  \subsection{Evaluation}

    \subsubsection{Metrics for Skill Evaluation}
    \label{sec:methods:metrics}
    Team performance is measured by four individual performance metrics:
    (1) success (i.e. the percentage of successful episodes),
    (2) the time to episode completion in seconds,
    (3) the number of simulation steps with collisions (Col) between instruments and organs, as well as between both instruments, and
    (4) the total path length (PL) of each instrument in millimeters.
    A shorter time to episode completion and shorter path length (i.e. efficiency of movement) are common metrics to evaluate surgical skill~\cite{hoveObjectiveAssessmentTechnical2010}.
    However, reckless behavior may also minimize the time and path length at the cost of collisions between instruments and organs, which may lead to bleeding or injury to the gallbladder with bile spillage during a real surgery.
    Thus, an additional performance metric monitors the number of simulation steps with collisions between objects.
    A high-quality episode can only be achieved if gripper and cauter cooperate effectively, lifting the gallbladder swiftly without losing grasp, and moving towards the target without colliding with other objects.

    \subsubsection{Evaluation of Artificial Teams}
    The artificial team was tested in the simulation environment for 500 episodes after the training was completed ($7$ million steps in the environment).
    During evaluation, the neural network parameters were frozen.

    \subsubsection{Evaluation of Human and Hybrid Teams}
    Initially, the three human teams outlined in Section~\ref{sec:methods:teams} interacted with the simulation environment until their respective learning curves plateaued.
    The team members decided when this was the case.
    After the learning phase, each team completed the task a total of ten (n~=~10) times, then filled out a NASA-TLX questionnaire~\cite{hart2006nasa}.
    The same procedure was repeated for the hybrid teams.

\section{RESULTS}
\label{sec:results}
  \subsection{Results of the Artificial Team}
\begin{figure}[tb]
       \parbox{\columnwidth}{
         \vspace{1.5mm}
         \begin{tikzpicture}
                \begin{axis}[
                        axis x line=bottom,
                        axis y line=middle,
                        axis line style={-},
                        x label style={at={(axis description cs:0.5,-0.15)},anchor=north},
                        y label style={at={(axis description cs:-0.1,0.5)},rotate=90,anchor=south},
                        xlabel={\small Total Environment Steps},
                        ylabel={\small Discounted Return},
                        ytick distance={5},
                        enlarge y limits=true,
                        width=0.95\columnwidth,
                        height=0.5\columnwidth,
                        grid = major,
                        grid style={dashed, gray!30},
                        legend style={at={(0.02,1)}, anchor=north west},
                        yticklabel style = {font=\scriptsize,xshift=0.5ex},
                        xticklabel style = {font=\scriptsize,yshift=0.5ex},
                        every x tick scale label/.style={at={(1,-0.1)},xshift=1pt,anchor=north,inner sep=0pt}
                     ]
                    \draw (axis description cs:0,1) -- (axis description cs:1,1);
                    \addplot[smooth, no markers, black, thick] table[x=steps,y=return,col sep=comma] {plots/data.csv};
                    \label{return_plot}
                \end{axis}

                \begin{axis}[
                        axis x line=bottom,
                        axis y line=right,
                        axis line style={-},
                        x label style={opacity=0},
                        y label style={at={(axis description cs:1.14,0.5)},rotate=0,anchor=south},
                        xlabel={},
                        ytick distance={25},
                        ylabel={\small Outcome [\%]},
                        enlarge y limits=true,
                        width=0.95\columnwidth, 
                        height=0.5\columnwidth,
                        legend style={fill opacity=1, draw opacity=1, text opacity=1, at={(1,0.15)}, anchor=south east, nodes={scale=0.8, transform shape}, draw=none},
                        yticklabel style = {font=\scriptsize,xshift=-0.5ex},
                        xticklabel style = {opacity=0},
                     ]
                    \addplot[smooth, no markers, OliveGreen, thick, dash dot] table[x=steps,y=won,col sep=comma] {plots/data.csv};
                    \addlegendentry{\small{Reached Goal}}
                    \addplot[smooth, no markers, red, thick, dashed] table[x=steps,y=lost,col sep=comma] {plots/data.csv};
                    \addlegendentry{\small{Lost Grasp}}
                    \addplot[smooth, no markers, blue, thick, dotted] table[x=steps,y=time,col sep=comma] {plots/data.csv};
                    \addlegendentry{\small{Ran Out of Time}}
                    \addlegendimage{/pgfplots/refstyle=return_plot}\addlegendentry{\small{Discounted Return}}
                    \begin{scope}[on background layer]
                        \fill[red,opacity=0.1] ({rel axis cs:0,0}) rectangle ({rel axis cs:0.2,1});
                        \fill[blue,opacity=0.1] ({rel axis cs:0.2,0}) rectangle ({rel axis cs:0.46,1});
                        \fill[OliveGreen,opacity=0.1] ({rel axis cs:0.46,0}) rectangle ({rel axis cs:1,1});
                    \end{scope}
                \end{axis}

            \end{tikzpicture}
       }
       \vspace{-0.2cm}
       \caption{Smoothed learning curve for one example training run. Discounted return and percentage of the three different outcomes are shown over steps in the learning environment.}
       \label{fig:learning_curve}
       \vspace{-0.5cm}
\end{figure}
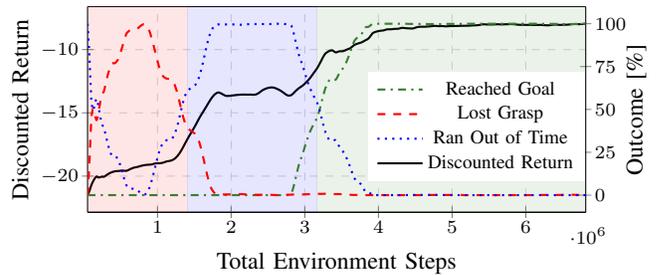

  The learning curve for one example training run is illustrated in Fig.~\ref{fig:learning_curve}.
  The discounted return increases with the number of training steps in the environment.
  The three possible outcomes separate the learning curve into distinct phases, illustrated by background color in Fig.~\ref{fig:learning_curve}.
  Losing the grasp on the gallbladder is the predominant outcome during early training stages.
  Over time, \textit{losing the grasp} becomes less frequent, increasing the discounted return.
  \textit{Running out of time} in an episode stays the most frequent outcome, before the agents start to reach the goal more consistently.
  Reaching the goal provides a strong learning signal, reducing the occurrence of \textit{running out of time}.

  The artificial team displays an average success rate of $99.8\pm0.04\%$ with $5.3\pm0.2$~\si{\second} to complete the episode.
  The path length reaches an average of $268\pm13$~\si{\mm} and $322\pm23$~\si{\mm} for gripper and cauter, respectively.
  Collisions are observed between cauter and liver in $0.01\pm0.18$~steps, between cauter and gallbladder in $0.3\pm1.65$~steps, and between gripper and liver in $0.12\pm0.49$~steps on average.
  No collisions were observed between instruments.

  \subsection{Results of the Human Teams}

      \begin{table*}
      \centering
        \vspace{1.5mm}
      \caption{Evaluation results for human and hybrid teams. Metrics from Section~\ref{sec:methods:metrics} are averaged over ten trials (n~=~10). The teams are abbreviated as illustrated in Fig.~\ref{fig:teams}. Simulation steps with collisions are observed between gripper and liver (Col-GL), cauter and liver (Col-CL), cauter and gallbladder (Col-CG), and between instruments (Col-II).}
      \begin{tabular*}{\textwidth}{ @{\extracolsep{\fill}} lrrrrrrrrr}
        \toprule
        Metric         & HuT1            & HuT2           & HuT3            & HyT1           & HyT2          & HyT3          & HyT4          & HyT5           & HyT6          \\
        \midrule
        Success [\%]   & $100 \pm 0$     & $100 \pm 0$    & $100 \pm 0$     & $100 \pm 0$    & $100 \pm 0$   & $100 \pm 0$   & $100 \pm 0$   & $100 \pm 0$    & $100 \pm 0$   \\
        Time [s]       & $19.8 \pm 2.6$  & $14.2 \pm 4.6$ & $13.2 \pm 3.3$  & $10.1 \pm 2.3$ & $5.7 \pm 0.4$ & $7.9 \pm 0.6$ & $5.7 \pm 0.3$ & $7.3 \pm 0.9$  & $5.1 \pm 0.4$ \\
        PL-G [mm]      & $295 \pm 12$    & $302 \pm 9$    & $315 \pm 24$    & $298 \pm 12$   & $257 \pm 5$   & $296 \pm 10$  & $266 \pm 9$   & $300 \pm 13$   & $261 \pm 8$   \\
        PL-C [mm]      & $297 \pm 46$    & $344 \pm 29$   & $341 \pm 56$    & $284 \pm 37$   & $347 \pm 29$  & $269 \pm 26$  & $359 \pm 14$  & $247\pm 37$    & $300 \pm 29$  \\
        Col-GL [steps] & $2.4 \pm 7.2$   & $0.0 \pm 0.0$  & $0.0 \pm 0.0$   & $0.0 \pm 0.0$  & $0.2 \pm 0.4$ & $0.1 \pm 0.3$ & $0.0 \pm 0.0$ & $0.0 \pm 0.0$  & $0.3 \pm 0.9$ \\
        Col-CL [steps] & $13.7 \pm 41.1$ & $3.4 \pm 10.2$ & $0.0 \pm 0.0$   & $0.0 \pm 0.0$  & $0.0 \pm 0.0$ & $0.0 \pm 0.0$ & $0.0 \pm 0.0$ & $7.8 \pm 23.4$ & $0.0 \pm 0.0$ \\
        Col-CG [steps] & $3.7 \pm 11.1$  & $1.5 \pm 4.2$  & $14.1 \pm 15.8$ & $1.1 \pm 3.3$  & $0.2 \pm 0.4$ & $1.7 \pm 4.0$ & $0.1 \pm 0.3$ & $0.0 \pm 0.0$  & $1.8 \pm 3.1$ \\
        Col-II [steps] & $0.0 \pm 0.0$   & $0.0 \pm 0.0$  & $0.0 \pm 0.0$   & $0.0 \pm 0.0$  & $0.0 \pm 0.0$ & $0.0 \pm 0.0$ & $0.0 \pm 0.0$ & $0.0 \pm 0.0$  & $0.0 \pm 0.0$ \\
        \bottomrule
      \end{tabular*}
      \label{tab:human_hybrid_team}
      \vspace{-0.5cm}
    \end{table*}

  The learning phases were completed after 10, 13, and 5 episodes for the first, second, and third human team, respectively.
  The results for the evaluation of the human teams are summarized in Table~\ref{tab:human_hybrid_team}.
  Each human team manages to solve the task consistently with a success rate of 100\%.
  HuT2 achieves the overall best performance in terms of simulation steps with collisions (Col-\{GL, CL, CG, II\}).
  The team with reversed roles (HuT3) exhibits a higher number of simulation steps with collisions between cauter and gallbladder (Col-CG), resulting in a shorter time to task completion (Time).
  The \textit{expert} (HuT1) performs similar or worse in each of the metrics except for instrument path lengths, with $\sim 4.5\%$ shorter paths for gripper (PL-G) and $\sim 13\%$ shorter paths for the cauter (PL-C).
  As expected, completion time suffers from sequential instrument control, whereas the other human teams may operate instruments simultaneously.

  The perceived workload as captured by the NASA-TLX questionnaire shows that controlling the cauter (\textit{resident}: 34.00, \textit{student}: 55.67) is perceived as twice as demanding as controlling the gripper (\textit{resident}: 17.00, \textit{student}: 37.00).
  The \textit{expert}'s weighted rating of 36.33 reflects control over both gripper and cauter and is approximately the average of the weighted ratings of the two other human teams.

  \subsection{Results of the Hybrid Teams}
  The learning phases were completed after 9, 3, 14, 7, 34, and 10 episodes for hybrid teams HyT1 to HyT6, respectively.
  The results for the evaluation of the hybrid teams are summarized in Table~\ref{tab:human_hybrid_team}.
  Similar to the human teams, each hybrid team manages to solve the task consistently with a success rate of 100\%.
  Teams with human control of the gripper and artificial control of the cauter (HyT\{2, 4, 6\}) require 55.6\%, 71.9\%, and 72.7\% less time to complete the task than the teams with reversed roles (HyT\{1, 3, 5\}).
  Path length, however, follows the opposite trend.
  The path length of the cauter is between 21.5\% and 33.5\% shorter when controlled by a human team member compared to the artificial agent.
  A similar but weaker trend is observed for the path length of the gripper, with 11.4\% to 16.0\% increase in path length.
  The number of simulation steps with collisions decreases noticeably in comparison to the human teams but increases in comparison to the artificial team.
  Only HyT5 exhibits a marked increase in collisions between cauter and liver (Col-CL) compared to the human teams.
  NASA-TLX questionnaires demonstrated a decreased overall perceived workload compared to the human teams.
  Control of the cauter, however, is still rated as approximately twice as demanding compared to control of the gripper.

\section{DISCUSSION}
\label{sec:discussion}

\subsection{Artificial Team}
The learning curve in Fig.~\ref{fig:learning_curve} exhibits three distinct phases.
First, the agents learn to navigate the environment safely by reducing the number of collisions and lost grasps.
Then, the agents explore the environment to learn the task itself, often running out of time.
Finally, the agents exploit the environment to maximize the return by reaching the target in a minimal number of simulation steps.

The multi-agent formulation of sub tasks in laparoscopic surgery is a promising approach to simplify several aspects of the learning problem.
For instance, it address the exponentially growing action space when controlling multiple instruments.
The action space in this work was of size $9$ per policy, with the total action space growing linearly with the number of instruments.
A single-agent policy would have required an action space of $9\times9$ to control the same number of instruments.
In addition, agent-specific rewards reduce the multi-agent credit assignment problem~\cite{foerster2018counterfactual} to learning only the temporal correlation between an agent's own actions and time-delayed rewards.
In this task, the gripper agent has to learn to lift the gallbladder and then remain idle and wait for the other policy to finish its task.
The cauter agent, on the other hand, has to react to the gripper moving the gallbladder out of the way, when approaching the target.
This coordinated movement is much more likely to be learned when rewards are assigned uniquely to each agent.
Preliminary experiments with a shared reward for all agents did not succeed in learning the task, presumably due to the inability to correctly correlate agent-specific actions to the observed shared reward.
Furthermore, separate policy models for each agent demonstrate a higher capacity to learn coordinated behavior compared to a shared policy model.
Initial experiments with a shared, multi-headed model for both agents resulted in satisfactory policies for the gripper agent, but very poor performance for the cauter agent.
The weights of a shared model may settle into a local optimum for gripper behavior in the first phase, such that the model cannot additionally learn a good cauter policy in the second phase.
This indicates that separate models per agent improve performance in tasks requiring sequential coordination between heterogeneous agents.

\subsection{Human and Hybrid Teams}
Controlling the cauter is perceived as more demanding, not only because more complex trajectories are required to solve the task, but also due to difficulties with depth perception in the simulation environment and controlling the surgical instruments through an Xbox controller.
The human teams reported having used the relative size of the cauter instrument and its shadow as features to estimate depth.
Lack of verbal communication in hybrid teams did not hinder performance.
This is likely because the human teams also completed the task silently, only speaking to offer suggestions in the learning phase.
The NASA-TLX results should only be interpreted as a qualitative trend since the sample size is very small and experiments were conducted on different days.

During the learning phase in hybrid teams, the artificial cauter agent sometimes could not recover from unexpected behavior by its human partner.
The artificial gripper agent, however, was highly robust against different skill levels of the human cauter operator.
This may be attributable to distribution shift between training and testing of the policies, suggesting that the gripper has experienced a wide distribution of cauter policies during training.
In contrast, the cauter was only able to significantly improve once the gripper already reached a certain performance level.
Since the task was relatively simple, the hybrid teams are still successful during testing, despite the distribution shift.
Robustness of policies that learn sequential behaviors is thus a key point in investigating assistive systems.
The notably shorter time to completion for the hybrid teams may be attributed to the agents learning to minimize episode length to maximize the discounted return.
The increased path length for the artificial agents, however, is a result of not considering path length or motion smoothness in the reward function.
\vspace{-0.5mm}
\subsection{Limitations}
Shaping the reward function as described in Section~\ref{sec:methods:spaces} is a major limitation of the presented method for learning decentralized policies.
The reward function requires substantial knowledge about the true state $s$ of the environment and involves fine tuning the component weights until the desired behavior is observed.
The proposed method cannot be transferred to a different environment without iteratively re-tuning the components.
The same re-tuning may also be required for the hyper parameters of \gls{ppo}.
Current and future work investigates algorithms that can cope with sparser reward functions in large state and action spaces to make the approach more transferable across tasks and increase robustness against variations in the environment.
Approaches from Inverse Reinforcement Learning may be utilized to avoid feature engineering complex reward functions by learning reward functions from expert demonstrations.

Furthermore, it may be argued that complex coordination of behaviors is not strictly required for the presented task since there is little variability among possible behaviors that solve it.
Future work will investigate whether the proposed methods scale well to more complex surgical tasks that require simultaneous coordination among instruments.

\gls{rl} in simulation environments is suited for exploring different approaches and algorithms for robotic surgery.
Transferring vision-based policies to real-world robotic systems, however, is restricted by the reality gap between simulated and real images.
Learning behaviors directly in the physical world, on the other hand, is restricted by low sample efficiency.
Domain translation~\cite{inoue_transfer_2017} and world model methods~\cite{hafner_dream_2020} may be key techniques to address these restrictions and bring \gls{rl} algorithms for control in robot-assisted surgery into the clinical reality.

Reducing the amount of required environment interactions and further improving robustness may be investigated by introducing expert knowledge into the reinforcement learning framework~\cite{kellerOpticalCoherenceTomographyGuided2020a}.
\vspace{-1.5mm}
\section{CONCLUSION}
\label{sec:conclusion}

This work demonstrates the first successful application of \gls{marl} in robot-assisted laparoscopic surgery.
Independent agents were trained to control laparoscopic instruments on a sub task from laparoscopic cholecystectomy in a soft body simulation.
In contrast to most examples from \gls{rl} literature, the task became easier to solve when formulated as a multi-agent, rather than a single-agent problem.
Evaluation of the agents in cooperation with human surgeons has demonstrated their suitability as autonomous assistants with rich potential for future integration into the clinic.

From a surgical perspective, the evaluated task is one of the main building blocks in surgery.
Applying tension to tissue and then dissecting it is important in numerous laparoscopic procedures.
Up until now, no actual dissection was performed, which should be a matter of future investigation.
Nevertheless, the mere application of \gls{marl} to robot-assisted surgery can be considered a major leap towards cognitive surgical robots that interact with the surgeon similar to a trained human assistant.

\bibliographystyle{IEEEtran.bst}
\bibliography{references}

\end{document}